%% file: emnlp2021.tex
\pdfoutput=1

\documentclass[11pt]{article}

\usepackage[]{emnlp2021}

\usepackage{times}
\usepackage{latexsym}

\usepackage[T1]{fontenc}

\usepackage[utf8]{inputenc}

\usepackage{times}
\usepackage{latexsym}
\usepackage{subcaption}
\usepackage{adjustbox}
\usepackage{multirow}
\usepackage{graphics}
\usepackage{microtype}
\usepackage{amsmath}
\usepackage{amssymb}
\usepackage{bm}
\usepackage{graphicx}
\usepackage{xcolor}
\usepackage{array}

\title{Condenser: a Pre-training Architecture for Dense Retrieval}

\author{Luyu Gao and Jamie Callan \\
  Language Technologies Institute \\
  Carnegie Mellon University \\
  \texttt{\{luyug, callan\}@cs.cmu.edu}  }

\begin{document}
\maketitle
\input{abstract}
\input{intro}
\input{related}
\input{method}
\input{experiment}

\input{analysis}
\input{conclusion}
\bibliography{anthology,acl2021}
\bibliographystyle{acl_natbib}

\clearpage
\appendix
\input{appendix}

\end{document}

%% file: abstract.tex
\begin{abstract}
Pre-trained Transformer language models~(LM) have become go-to text representation encoders. Prior research fine-tunes deep LMs to encode text sequences such as sentences and passages into single dense vector representations for efficient text comparison and retrieval. However, dense encoders require a lot of data and sophisticated techniques to effectively train and suffer in low data situations.
This paper finds a key reason is that standard LMs' internal attention structure is not ready-to-use for dense encoders, which needs to aggregate text information into the dense representation. We propose to pre-train towards dense encoder with a novel Transformer architecture, Condenser, where LM prediction CONditions on DENSE Representation. Our experiments show Condenser improves over standard LM by large margins on various text retrieval and similarity tasks.\footnote{Code available at \url{https://github.com/luyug/Condenser}}

\end{abstract}

%% file: intro.tex
\section{Introduction}
Language model~(LM) pre-training has been very effective in learning text encoders that can be fine-tuned for many downstream tasks~\cite{peters-etal-2018-deep,devlin-etal-2019-bert}. Deep bidirectional Transformer encoder~\cite{Vaswani2017AttentionIA} LMs like BERT~\cite{devlin-etal-2019-bert} are the state-of-the-art. Recent works fine-tune the CLS token to encode input text sequence into a single vector representation~\cite{lee-etal-2019-latent,chang2020pretraining,karpukhin-etal-2020-dense}. The resulting model is referred to as dense encoder or bi-encoder. Fine-tuning associates with vector similarities some practical semantics, e.g., textual similarity or relevance, and therefore the vectors can be used for efficient text comparison or retrieval by inner product. Despite their efficiency, bi-encoders are hard to train. Even with sufficient data, bi-encoders still require carefully designed sophisticated methods to train effectively~\cite{xiong2021approximate,Qu2020RocketQAAO,Lin2020DistillingDR}. They can also take big performance hits in low data situations~\cite{karpukhin-etal-2020-dense,thakur2020augmented,chang2020pretraining}.
Another common use of deep LM is cross-encoder, pass compared text pair directly in and use attention overall tokens to do prediction. In contrast to bi-encoder, cross encoder trains easier and is effective in low data for similarity and ranking tasks~\cite{devlin-etal-2019-bert,Yang2019XLNetGA}.

Based on the same LM, however, bi-encoder and cross encoder have similar language understanding capabilities. To explain the difficulty in training bi-encoder not seen in cross-encoder, we look into the internal structure of pre-trained LM. We find LM like BERT directly out of pre-training has a non-optimal attention structure. In particular, they were not trained to aggregate sophisticated information into a single dense representation. We term effort during fine-tuning to adjust the LM internal activation to channel its knowledge out for the target task, \emph{structural readiness}. We argue bi-encoder fine-tuning is inefficient due to the lacking structural readiness. Many updates are used to adjust model attention structure than learn good representation.

Based on our observations, we propose to address structural readiness during pre-training. We introduce a novel Transformer pre-training architecture, Condenser, which establishes structural readiness by doing LM pre-training actively CONdition on DENSE Representation. Unlike previous works that pre-train towards a particular task, Condenser pre-trains towards the bi-encoder structure. Our results show the importance of structural readiness. We experiment with sentence similarity tasks, and retrieval for question answering and web search. We find under low data setups, with identical test time architecture, Condenser yields sizable improvement over standard LM and shows comparable performance to strong task-specific pre-trained models. With large training data, we find Condenser retriever optimize more easily, outperforming previous models trained with complicated techniques with a single round of negative mining.



%% file: related.tex
\section{Related Work}
\paragraph{Transformer Bi-encoder} LM pre-training followed by task fine-tuning has become one important paradigm in NLP~\cite{howard-ruder-2018-universal}. SOTA models adopt the Transformer architecture~\cite{devlin-etal-2019-bert,Liu2019RoBERTaAR,Yang2019XLNetGA,Lan2020ALBERTAL}. 
One challenge for applying deep Transformer is their computation cost when used to retrieve text from large collections. Motivated by this, \citet{reimers-gurevych-2019-sentence} propose SBERT which trains bi-encoder from BERT and uses vector product for efficient sentence similarity comparison. Transformer bi-encoders were soon also adopted as dense retriever~\cite{lee-etal-2019-latent,chang2020pretraining,karpukhin-etal-2020-dense, CLEAR}. 

\paragraph{Dense Retrieval} Dense retrieval compares encoded query vectors with corpus document vectors using inner product. While there are works on efficient cross-encoder~\cite{gao-etal-2020-modularized, MacAvaney2020EfficientDR}, such models are still too costly for full corpus retrieval. By pre-encoding the corpus into MIPS~\cite{JDH17,avq_2020} index, retrieval can run online with millisecond-level latency. An alternative is the recently proposed contextualized sparse retrieval model~\cite{gao-etal-2021-coil}. In comparison, dense retrieval is easier to use and backed by more matured software like FAISS~\cite{JDH17}.

\paragraph{Pre-train Bi-encoder} 
\citet{lee-etal-2019-latent} are among the first to show the effectiveness of Transformer bi-encoder for dense retrieval. They proposed to further pre-train BERT with Inverse Cloze Task (ICT). ICT uses pair of passage segment and full passage as pseudo training pair. \citet{chang2020pretraining} find  ICT and other related tasks are ``key ingredients'' for strong bi-encoders. Their results also show that models without pre-training fail to produce useful retrieval results under low data setups. \citet{Guu2020REALMRL} propose to pre-train retriever and reader together for end-to-end QA system. The aforementioned methods are specialized \emph{task specific} solutions for improving bi-encoder training based on contrastive loss. This paper provides an \emph{explanation} for the learning issue and presents an architecture that establishes a universal solution using general language model pre-training. We also note that language model and contrastive pre-training are orthogonal ideas. In a follow-up work, we show further improved performance adding contrastive learning to Condenser language model pre-training~\cite{gao2021unsupervised}.

\paragraph{Effective Dense Retriever}  \citet{karpukhin-etal-2020-dense} found carefully fine-tuning BERT can produce better results than earlier pre-trained dense retrieval systems. To further improve the end performance of dense retrievers, later works look into better fine-tuning techniques. Using a learned retriever to mine hard negatives and re-train another retriever with them was found helpful~\cite{karpukhin-etal-2020-dense,Qu2020RocketQAAO}. ANCE~\cite{xiong2021approximate} actively mines hard negatives once after an interval during training to prevent diminishing gradients. It allocates extra resources to update and retrieve from the corpus retrieval index repetitively. \cite{CLEAR} proposed to jointly learn a pair of dense and sparse systems to mitigate the capacity issue with low dimension dense vectors. Beyond fine-tuning, using more sophisticated knowledge distillation loss to learn bi-encoders based on soft labels has also been found useful~\cite{chen-etal-2020-dipair,Lin2020DistillingDR}. They first learn a teacher model and use its predictions at training time to optimize the dense retriever. These works all aim at producing better gradient updates during training, while Condenser aims at better initializing the model. We will also show the combined improvement of Condenser and hard negatives in experiments. Another line of works question the capacity of single vector representation and propose to use multi-vector representation~\cite{Luan2020SparseDA}. Capacity defines the performance upper bound and is one other issue than training~(optimization), i.e. how to reach the upper bound.

\paragraph{Sentence Representation}We'd also like to make a distinction from works in universal sentence representation and encoder~\cite{kiros2015skipthought,conneau-etal-2017-supervised,cer2018universal}. They are \emph{feature-based} methods rather than fine-tuning~\cite{Houlsby2019ParameterEfficientTL}. In evaluation, they focus on using the learned embedding as universal features for a wide range of tasks~\cite{Conneau2018SentEvalAE}. This paper considers task-specific fine-tuning of the \emph{entire} model and focuses on the target task performance. 

%% file: method.tex


\section{Method}
\label{sec:method}
This section discusses the motivation behind Condenser, its design, and its pre-training procedure.
\subsection{Preliminaries}
\paragraph{Transformer Encoder}
Many recent state-of-the-art deep LM adopts the architecture of Transformer encoder. It takes in a text sequence, embed it and pass it through a stack of $L$ self-attentive Transformer blocks. Formally, given input text $x = [x_1, x_2, ...]$, we can write iteratively,
\begin{align}
    h^0 &= \text{Embed}(x)\\
    h^{l} &= \text{Transformer}_l(h^{l-1})
\end{align}
Intuitively, Transformer blocks refine each token's representation conditioning on all tokens in the sequence to effectively embed them. 
\paragraph{Transformer LM Pre-training}
Many successful Transformer Encoder LMs such as BERT are trained with masked language model~(MLM) task. MLM masks out a subset of input tokens and requires the model to predict them. For a masked out token $x_i$ at position $i$, its corresponding final representation $h_i^L$ is used to predict the actual $x_i$. Training uses a cross-entropy loss,
\begin{equation}
    \mathcal{L}_\text{mlm} = \sum_{i \in \text{masked}} \text{CrossEntropy}(W h^L_i, x_i)
\end{equation}
A special token, typically referred to as CLS is prepended and encoded with the rest of the text.
\begin{align}
    [h_{cls}^0;h^0] &= \text{Embed}([\text{CLS}; x])\\
    [h_{cls}^l;h^l] &= \text{TF}_l([h_{cls}^{l-1};h^{l-1}])
\end{align}
Some models train CLS explicitly during pre-training, notably BERT's next sentence prediction~(NSP; \citet{devlin-etal-2019-bert}), while others implicitly \cite{Yang2019XLNetGA,Liu2019RoBERTaAR}.


\subsection{Issues with Transformer Encoder}
Recall in Transformers, all tokens, including the CLS, receive information of other tokens in the sequence only with attention. Attention patterns, therefore, define how effective CLS can aggregate information. To understand the attentive behaviors of CLS, we borrow analysis of BERT from \citet{Clark2019WhatDB}: 1)~in most middle layers, the CLS token has similar attention patterns as other text tokens and is not attended by other tokens, 2)~until the last layer, CLS has unique broad attention over the entire sequence to perform NSP task. In other words, the CLS token remains dormant in many middle layers and reactivates only in the last round of attention. We argue that an effective bi-encoder should actively aggregate information of different granularity from the entire sentence through all layers, and this structure in standard pre-trained LM is not immediately ready for fine-tuning. We will verify this claim with experiments in \autoref{sec:experiment} and with quantitative analysis of attention of BERT, ICT, and the proposed Condenser in \autoref{sec:analysis}.

\begin{figure}[h]
  \centering
  \includegraphics[width=0.47\textwidth]{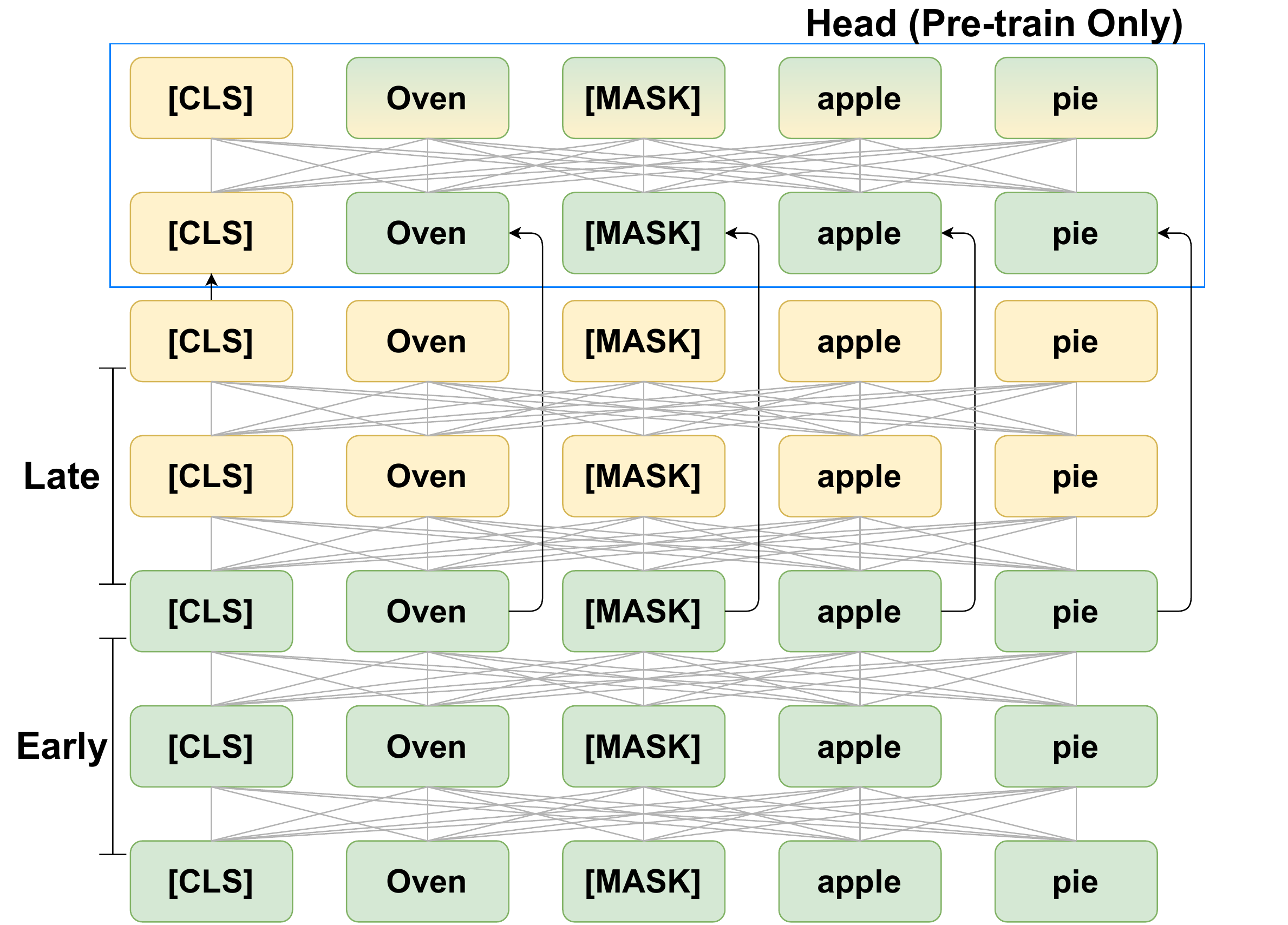}
  \caption{Condenser: We show 2 early and 2 late backbone layers here, in our experiments each have 6 layers. Condenser Head is dropped during fine-tuning.}
  \label{fig: model}
\end{figure}

\subsection{Condenser}
\label{subsec:condenser}
Building upon Transformer encoder LMs, which conditions on left and right context~\cite{devlin-etal-2019-bert}, we present bi-encoder pre-training architecture Condenser, which CONdition actively on DENSE Representation in LM pre-training.  
\paragraph{Model Design} 
Like Transformer Encoder, Condenser is parametrized into a stack of Transformer blocks, shown in \autoref{fig: model}. We divide them into three groups, $L^e$ early encoder backbone layers, $L^l$ late encoder backbone layers, and $L^h$ Condenser head Layers.
Inputs is first encoded by backbone,
\begin{align}
    &[h_{cls}^{early};h^{early}] = \text{Encoder}_\text{early}([h_{cls}^0;h^0]) \label{eq:cd-input} \\ 
    &[h_{cls}^{late};h^{late}] = \text{Encoder}_\text{late}([h_{cls}^{early};h^{early}]) 
\end{align}
\paragraph{Condenser Head} The critical design is that we put a short circuit from early output to the head, which takes in a pair of \emph{late-early} representations,
\begin{equation}
    [h_{cls}^{cd};h^{cd}] = \text{Condenser}_\text{head}([h_{cls}^{late};h^{early}])
\end{equation}
We train with MLM loss with the head's output,
\begin{equation}
    \mathcal{L}_\text{mlm} = \sum_{i \in \text{masked}} \text{CrossEntropy}(W h^{cd}_i, x_i)
\end{equation}
We follow the masking scheme in \citet{devlin-etal-2019-bert} to combat train test difference.

Within Condenser, the late encoder backbone can further refine the token representations but can only pass new information through $h_{cls}^{late}$, the late CLS. The late CLS representation is therefore required to aggregate newly generated information later in the backbone, and the head can then \emph{condition} on late CLS to make LM predictions. 
Meanwhile, skip connecting the early layers, we remove the burden of encoding local information and the syntactic structure of input text, focusing CLS on the global meaning of the input text. Layer numbers $L^e$ and $L^l$ control this separation of information.

Architecture of Condenser is inspired by Funnel Transformer~\cite{Dai2020FunnelTransformerFO}, which itself is inspired by U-net~\cite{RonnebergerFB15} from computer vision.  Funnel Transformer reduces sequence length by a factor of 4 during forward and uses a 2-layer Transformer to decode the length compressed sequence onto a skip-connected full-length representation. Funnel Transformer was designed to speed up pre-training while our Condenser learns dense information aggregation. 

\paragraph{Fine-tuning} The Condenser head is a pre-train time component and is dropped during fine-tuning. Fine-tuning trains the late CLS $h_{cls}^{late}$ and backpropagate gradient into the backbone. In other words, a Condenser reduces to its encoder backbone, or effectively becomes a Transformer encoder for fine-tuning; the head is only used to guide pre-training. During fine-tuning, Condenser has an identical capacity as a similarly structured Transformer. In practice, Condenser can be a drop-in weight replacement for a typical Transformer LM like BERT.

\subsection{Condenser from Transformer Encoder}
\label{sec:method-from-tf}
In this paper, we opted to initialize Condenser with pre-trained Transformer LM weight. This accommodates our compute budget, avoiding the huge cost of pre-training from scratch. This also gives us a direct comparison to the original LM.
Given a pre-trained LM, we initialize the entire Condenser backbone with its weights and randomly initialize the head. To prevent gradient back propagated from the random head from corrupting backbone weights, we place a semantic constraint by performing MLM also with backbone late outputs,
\begin{equation}
    \mathcal{L}_\text{mlm}^c = \sum_{i \in \text{masked}} \text{CrossEntropy}(W h^{late}_i, x_i)
\end{equation}
The intuition behind this constraint is that encoding per-token representations $h^{late}$ and sequence representation $h^{late}_{cls}$ share similar mechanism and will not interfere with each other. As a result, $h^{late}$ can still be used for LM prediction. The full loss is then defined as a sum of two MLM losses,
\begin{equation}
    \mathcal{L} = \mathcal{L}_\text{mlm} + \mathcal{L}_\text{mlm}^c
\end{equation}
The output projection matrix $W$ is shared between the two MLM losses to reduces the total number of parameters and memory usage.

%% file: experiment.tex
\section{Experiments}
\label{sec:experiment}
In this section, we first describe details on how to pre-train Condenser from BERT. Our fine-tuning experiments then look into the impacts of Condenser under low and high data setup. To evaluate low data, we sample smaller training sets similar to \citet{chang2020pretraining}, by sub-sampling the original train set. We keep dev/test sets unchanged across runs for direct comparison. We first validate our model with short sentence level tasks, then evaluate retrieval in open question answering and web search tasks following prior works~\cite{chang2020pretraining,xiong2021approximate}. We will examine how swapping original BERT with Condenser improves performance, and how the improvements compare to various improved training techniques.


\subsection{Pre-training}
We initialize Condenser backbone layers from the popular 12-layer BERT base and only a 2-layer head from scratch. Pre-training runs with procedures described in \autoref{sec:method-from-tf}. We use an equal split, 6 early layers, and 6 late layers. We pre-train over the same data as BERT: English Wikipedia and the BookCorpus. This makes sure BERT and Condenser differ only in architecture for direct comparison.
We train for 8 epochs, with AdamW, learning rate of 1e-4 and a linear schedule with warmup ratio 0.1. 
Due to compute budget limit, we were not able to tune the optimal layer split, head size or train hyperparameters, but leave that to future work. We train on 4 RTX 2080ti with gradient accumulation. The procedure takes roughly a week to finish. After pre-training, we discard the Condenser head, resulting in a Transformer model of the \emph{same} architecture as BERT. \emph{All} fine-tuning experiments share this \emph{single} pre-trained weight. 
\subsection{Sentence Similarity}

\paragraph{Dataset} We use two supervised data sets: Semantic Textual Similarity Benchmark(STS-b; \citet{cer-etal-2017-semeval}) and Wikipedia Section Distinction~\cite{ein-dor-etal-2018-learning} adopted in \citet{reimers-gurevych-2019-sentence}. The former is a standard sentence similarity task from GLUE~\cite{wang-etal-2018-glue} with a small training set~($\sim$6K). The latter is large($\sim$1.8M) and has an interesting objective, to determine if a pair of sentences are from the same Wikipedia section, very similar to the BERT NSP task. \citet{Lan2020ALBERTAL} argue NSP learns exactly topical consistency on the training corpus, i.e. Wikipedia. In other words, NSP is a close pre-training, if not training, task for Wiki Section Distinction.
We report test set Spearman correlation for STS-b and accuracy for Wiki Section Distinction.

\paragraph{Compared Systems} We compare with standard BERT and on STS-b, with BERT pre-trained with multiple NLI data sets with a popular carefully crafted 3-way loss~\cite{conneau-etal-2017-supervised} from \citet{reimers-gurevych-2019-sentence}\footnote{These models are referred to as SBERT in the original paper. We use BERT for consistency with later discussions.}. Non-BERT baselines are also borrowed from it.

\paragraph{Implementation} We use the sentence transformer software and train STS-b with MSE regression loss and Wiki Section with triplet loss~\cite{reimers-gurevych-2019-sentence}. 
The training follows the authors' hyper-parameter settings.

\paragraph{Results} \autoref{tab:sts-b} shows performance on STS-b with various train sizes. NLI pre-trained BERT and Condenser consistently outperform BERT and has a much larger margin with smaller train sizes. 
Also, with only 500 training pairs, they outperform the best Universal Sentence Encoder(USE) baseline.

For Wiki Section, in \autoref{tab:wiki-sec} we observe almost identical results among BERT and Condenser models, which outperform pre-BERT baselines. Meanwhile, even when training size is as small as 1K, we observe only about 10\% accuracy drop than training with all data. Without training with the NSP task, Condenser remains effective.
\begin{table}[h]
\centering
\begin{tabular}{ l || c c c }
\hline \hline
\multicolumn{4}{c}{\textbf{STS-b}}\\
\hline
\textbf{Model} & \multicolumn{3}{c}{\textbf{Spearman}} \\
\hline
GloVe  & \multicolumn{3}{c}{58.0}  \\
Infersent  & \multicolumn{3}{c}{68.0}  \\
USE  & \multicolumn{3}{c}{74.9}  \\
\hline
 \textbf{Train Size} & 500 & 1K & FULL  \\
 \hline
BERT & 68.6 & 71.4 & 82.5   \\
BERT + NLI & 76.4 & 76.8 & 84.7 \\
Condenser & \textbf{76.6} & \textbf{77.8} & \textbf{85.6}  \\
\hline \hline 
\end{tabular}
\caption{\textbf{STS-b}: Spearman correlation on Test Set. 
}
\label{tab:sts-b}
\end{table}

\begin{table}[h]
\centering
\begin{tabular}{ l || c c c }
\hline \hline
\multicolumn{4}{c}{\textbf{Wikipedia Section Distinction}}\\
\hline
\textbf{Model} & \multicolumn{3}{c}{\textbf{Accuracy}} \\
\hline
skip-thoughts  & \multicolumn{3}{c}{0.62}  \\
\hline
 \textbf{Train Size} & 1K & 10K & FULL  \\
 \hline
BiLSTM & n.a. & n.a. & 0.74  \\
BERT  & 0.72 & 0.75 & 0.80   \\
Condenser  & 0.73 & 0.76 & 0.80  \\
\hline \hline 
\end{tabular}
\caption{\textbf{Wiki Section}: Accuracy on Test Set.}
\label{tab:wiki-sec}
\vspace{-0.4cm}
\end{table}


\begin{table*}[t]
\centering
\scalebox{0.95}{
\begin{tabular}{ l || c c c | c c c | c c c | c c c}
\hline \hline
 & \multicolumn{6}{c|}{\textbf{Natural Question}} & \multicolumn{6}{c}{\textbf{Trivia QA}}  \\
Model & \multicolumn{3}{c}{Top-20} & \multicolumn{3}{c|}{Top-100} & \multicolumn{3}{c}{Top-20} & \multicolumn{3}{c}{Top-100}  \\
\hline
BM25 & \multicolumn{3}{c|}{59.1} & \multicolumn{3}{c|}{73.7} & \multicolumn{3}{c|}{66.9} & \multicolumn{3}{c}{76.7}\\
\hline
 \textbf{Train Size} & 1K & 10K & FULL & 1K & 10K & FULL & 1K & 10K & FULL & 1K & 10K & FULL\\
 \hline
BERT & 66.6 &  75.9	& 78.4 & 79.4 & 84.6 & 85.4 & 68.0 & 75.0 & 79.3 & 78.7 & 82.3 & 84.9 \\
ICT & \textbf{72.9} & \textbf{78.4} & \textbf{80.9} & \textbf{83.7} & \textbf{85.9} & \textbf{87.4} & 73.4 & 77.9 & 79.7 & \textbf{82.3} & 84.8 & 85.3 \\
Condenser & 72.7 & \textbf{78.3} & 80.1 & 82.5  & \textbf{85.8} & 86.8 & \textbf{74.3} & \textbf{78.9} & \textbf{81.0} & \textbf{82.2} & \textbf{85.2} & \textbf{86.1} \\
\hline \hline 
\end{tabular}
}
\caption{Low data: Results on Natual Question and Triavia QA measured by Top-20/100 Hits. Models in this table are all trained with BM25 negatives. Results within 0.1 difference with the best are marked bold.}
\label{tab:openqa}
\end{table*}

\subsection{Retrieval for Open QA}
In this section, we test bi-encoders with open QA passage retrieval experiments~\cite{chang2020pretraining,karpukhin-etal-2020-dense}.
Compared to the sentence level task, search tasks explicitly use the learned structure of the embedding space, where similarity corresponds to the relevance between a pair of query, passage. 
We adopt the DPR~\cite{karpukhin-etal-2020-dense} setup, fine-tune LM with a contrastive loss in training, computing for query $q$, the negative log likelihood of a positive document $d^+$ against a set of negatives $\{d^-_1, d^-_2, .. d^-_l ..\}$. 

\begin{equation}
\mathcal{L} = -\log \frac{\exp(s(q, d^+))}{ \exp(s(q, d^+)) + \underset{l}{\sum} \exp(s(q, d^-_l)) }
\end{equation}
Negatives can come from various sources: random, top BM25, hard negatives, or sophisticatedly sampled like ANCE. We conduct low data experiments with BM25 negatives to save compute and use mined hard negatives~(HN) in full train experiments.

\paragraph{Dataset} We use two query sets, Natural Question(NQ; \citet{kwiatkowski-etal-2019-natural}) and Trivia QA(TQA; \citet{joshi-etal-2017-triviaqa}), as well as the Wikipedia corpus cleaned up and released with DPR. NQ contains questions from Google search and TQA contains a set of trivia questions. Both NQ and TQA have about 60K training data post-processing. We refer readers to \citet{karpukhin-etal-2020-dense} for details. We adopt DPR evaluation metrics, report test set hit accuracy of Top-20/100.

\paragraph{Compared Systems} For low data experiments, we compare BERT, ICT, and Condenser. We attempted to train ICT on our hardware for direct comparison but found the end result bad, due to the small batch size. We instead use ICT released by \citet{lee-etal-2019-latent} trained with 4096 batch size from BERT for more informative comparison.\footnote{A detailed discussion of this choice of ICT is in \ref{app:ict}} For full train, we compare with lexical systems BM25 and GAR~\cite{mao2020generationaugmented} and dense systems DPR~(BERT), DPR with HN and ANCE. GAR uses a learned deep LM BART~\cite{lewis-etal-2020-bart} to expand queries. ANCE uses asynchronous corpus index update~\cite{Guu2020REALMRL} to do multiple rounds of hard negative mining during training. We also compare with RocketQA~\cite{Qu2020RocketQAAO}, which is trained with an optimized fine-tuning pipeline that combines hard negative, large~(1024) batch, supervision from cross-encoder, and external data.

\paragraph{Implementation} We train Condenser systems using the DPR hyper-parameter setting. We use a single RTX 2080ti and employ the gradient cache technique~\cite{gao-etal-2021-scaling} implemented in the GC-DPR toolkit\footnote{\url{https://github.com/luyug/GC-DPR}} to perform large batch training with the GPU's limited memory.
As DPR only released Natural Question hard negatives, we use theirs on Natural Question and mine our own with a Condenser retriever on TriviaQA.

\paragraph{Results} In \autoref{tab:openqa}, we record test set performance for NQ and TQA with low data. We observe ICT and Condenser both outperform vanilla BERT, by an especially large margin at 1K training size, dropping less than 10\% compared to full-size training for Top-20 Hit and less than 5\% for Top-100. The improvement is more significant when considering the gain over unsupervised BM25. ICT and Condenser show comparable performance, with ICT slightly better on NQ and Condenser on TQA. This also agrees with results from \citet{lee-etal-2019-latent}, that ICT specializes in NQ. The results suggest general LM-trained Condenser can be an effective alternative to task-specific pre-trained model ICT.

In \autoref{tab:qa-full}, we compare Condenser trained with full training data with other systems. On NQ, dense retrievers all yield better performance than lexical retrievers, especially those that use hard negatives. We see Condenser performs the best for Top-20 and is within 0.1 to RocketQA for Top-100, without requiring the sophisticated and costly training pipeline. On TQA, we see GAR, lexical with deep LM query expansion, perform better than all dense systems other than Condenser. This suggests TQA may require granular term-level signals hard to capture for dense retrievers. Nevertheless, we find Condenser can still capture these signals and perform better than all other lexical and dense systems.

\begin{table}[h]
\centering
\begin{tabular}{ l || c c | c c }
\hline \hline
 & \multicolumn{2}{c|}{\textbf{NQ}} & \multicolumn{2}{c}{\textbf{TQA}} \\
Model & \multicolumn{2}{c|}{Top-20/100} & \multicolumn{2}{c}{Top-20/100} \\
\hline
BM25 & 59.1 & 73.7 & 66.9 & 76.7 \\
GAR & 74.4 & 85.3 & 80.4 & 85.7 \\
\hline
DPR & 78.4 & 85.4 & 79.3 & 84.9   \\
DPR + HN & 81.3 & 87.3 & 80.7 & 85.8 \\
ANCE & 81.9 & 87.5 &  80.3 & 85.3 \\
RocketQA &  82.7 & \textbf{88.5} & n.a. & n.a.\\
\hline
Condenser & \textbf{83.2} & \textbf{88.4} & \textbf{81.9} & \textbf{86.2} \\
\hline \hline 
\end{tabular}
\caption{Full train for Natural Question and Trivia QA. Results not available are denoted `n.a.' Results within 0.1 difference with the best are marked bold.}
\label{tab:qa-full}
\vspace{-0.6cm}
\end{table}


\begin{table*}[t]
\centering
\begin{tabular}{ p{6em} || c c c | c c c | c c c }
\hline \hline
 & \multicolumn{6}{c|}{MS-MARCO Dev} & \multicolumn{3}{c}{DL2019}  \\
Model & \multicolumn{3}{>{\centering\arraybackslash}p{9em}}{MRR@10} & \multicolumn{3}{>{\centering\arraybackslash}p{9em}|}{Recall@1000} & \multicolumn{3}{>{\centering\arraybackslash}p{9em}}{NDCG@10} \\
\hline
BM25 & \multicolumn{3}{c|}{0.184} & \multicolumn{3}{c|}{0.853} & \multicolumn{3}{c}{0.506} \\
\hline
 \textbf{Train Size} & 1K & 10K & FULL & 1K & 10K & FULL & 1K & 10K & FULL \\
 \hline
BERT & 0.156 & 0.228 & 0.309 & 0.786 & 0.878 & 0.938 & 0.424 & 0.555 & 0.612 \\
ICT & 0.175 & 0.251 & 0.307 & 0.847 & 0.905 & 0.945 & 0.519 & 0.585 & 0.624 \\
Condenser & \textbf{0.192} & \textbf{0.258} & \textbf{0.338}  & \textbf{0.852} & \textbf{0.914} & \textbf{0.961} & \textbf{0.530}	& \textbf{0.591} & \textbf{0.648} \\
\hline \hline 
\end{tabular}
\caption{Low data: Performacne is measured by MRR@10, Recall@1k. Models in this table are all trained with BM25 negatives.}
\label{tab:marco}
\vspace{-0.5cm}
\end{table*}


\subsection{Retrieval for Web Search} 
In this section, we examine how Condenser retriever performs on web search tasks. The setup is similar to open QA. One issue with web search data sets is that they are noisier, containing a large number of false negatives~\cite{Qu2020RocketQAAO}. We investigate if Condenser can help resist such noise. As passage retrieval is the focus of the paper, we defer discussion of long document retrieval to \ref{app:mdoc}.

\paragraph{Dataset} We use the MS-MARCO passage ranking dataset~\cite{bajaj2018ms}, which is constructed from Bing's search query logs and web documents retrieved by Bing. The training set has about 0.5M queries. We use corpus pre-processed and released with RocketQA. 
We evaluate on two query sets: MS-MARCO Dev\footnote{The test set was hidden; MS-MARCO organizers discourage multi submissions but recommend studies over Dev set.} and TREC DL2019 queries. We report on Dev official metrics MRR@10 and Recall@1k, and report on DL2019 NDCG@10. 


\paragraph{Implementation}  We train with the contrastive loss with a learning rate of 5e-6 for 3 epochs on a RTX2080ti. We pair each query with 8 passages as \citet{Luan2020SparseDA} and use a total batch of 64 passages. Low data experiments use BM25 negatives and full data experiments use hard negatives mined with BM25 negative trained Condenser.

\paragraph{Compared Systems} For low data settings, we again compare BERT, ICT, and Condenser. Here, all the three are not trained on the MS-MARCO corpus; we examine their generalization capability. For full training setup, we compare with lexical system BM25, deep LM augmented lexical systems DeepCT~\cite{DeepCT} and DocT5Qry~\cite{docTTTTTquery}, and dense systems, ANCE, TCT~\cite{Lin2020DistillingDR} and ME-BERT~\cite{Luan2020SparseDA}. TCT also aims at improving training like ANCE, but by replacing contrastive loss fine-tuning with knowledge distillation. ME-BERT uses BERT large variant as encoder, three times larger than LMs used in other systems, and represents passage with multiple vectors. It gets higher encoder and embedding capacity but has higher costs in train, inference, and retrieval. Since the full RocketQA system uses data external to MS-MARCO, for a fair comparison, we include the variant without external data in the main result \autoref{tab:marco-full} and separately compare Condenser with all RocketQA variants in \autoref{tab:rocket}.

\begin{table}[h]
\centering
\scalebox{0.9}{
\begin{tabular}{ l || c  c | c }
\hline \hline
 & \multicolumn{2}{c|}{MS-MARCO Dev} & DL2019 \\
Model & \multicolumn{1}{c}{MRR@10} & R@1K & NDCG@10 \\
\hline
BM25 & 0.189 & 0.853 & 0.506 \\
DeepCT & 0.243 & 0.909 & 0.572\\
DocT5Qry & 0.278 & 0.945 & 0.642\\
\hline
BERT & 0.309 & 0.938 & 0.612    \\
BERT + HN & 0.334 & 0.955 & 0.656 \\
ME-BERT & 0.334 & n.a. & 0.687 \\
ANCE & 0.330 & 0.959 &  0.648  \\
TCT & 0.335 & 0.964 & 0.670\\
RocketQA* & 0.364 & n.a. & n.a. \\
\hline
Condenser & \textbf{0.366} & \textbf{0.974} & \textbf{0.698} \\
\hline \hline 
\end{tabular}
}
\caption{Full train setup on MS-MARCO. Results not available are denoted `n.a.' *: RocketQA variant here is not trained with external data.}
\label{tab:marco-full}
\end{table}

\begin{figure*}[t!]
\centering
    \begin{subfigure}[t]{0.33\textwidth}
        \centering
        \includegraphics[width=\textwidth]{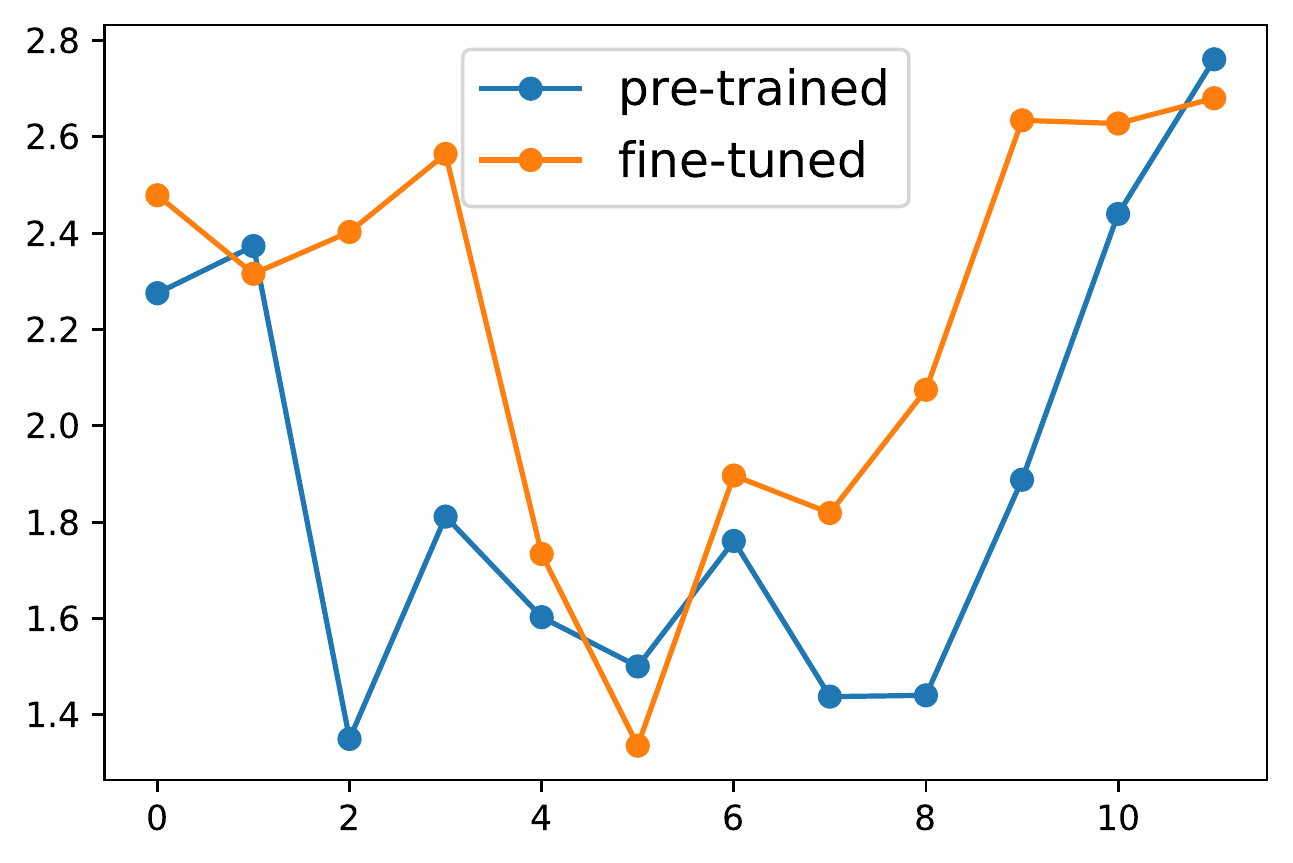}
        \caption{BERT}
        \label{fig:attn-bert}
    \end{subfigure}~
    \begin{subfigure}[t]{0.33\textwidth}
        \centering
        \includegraphics[width=\textwidth]{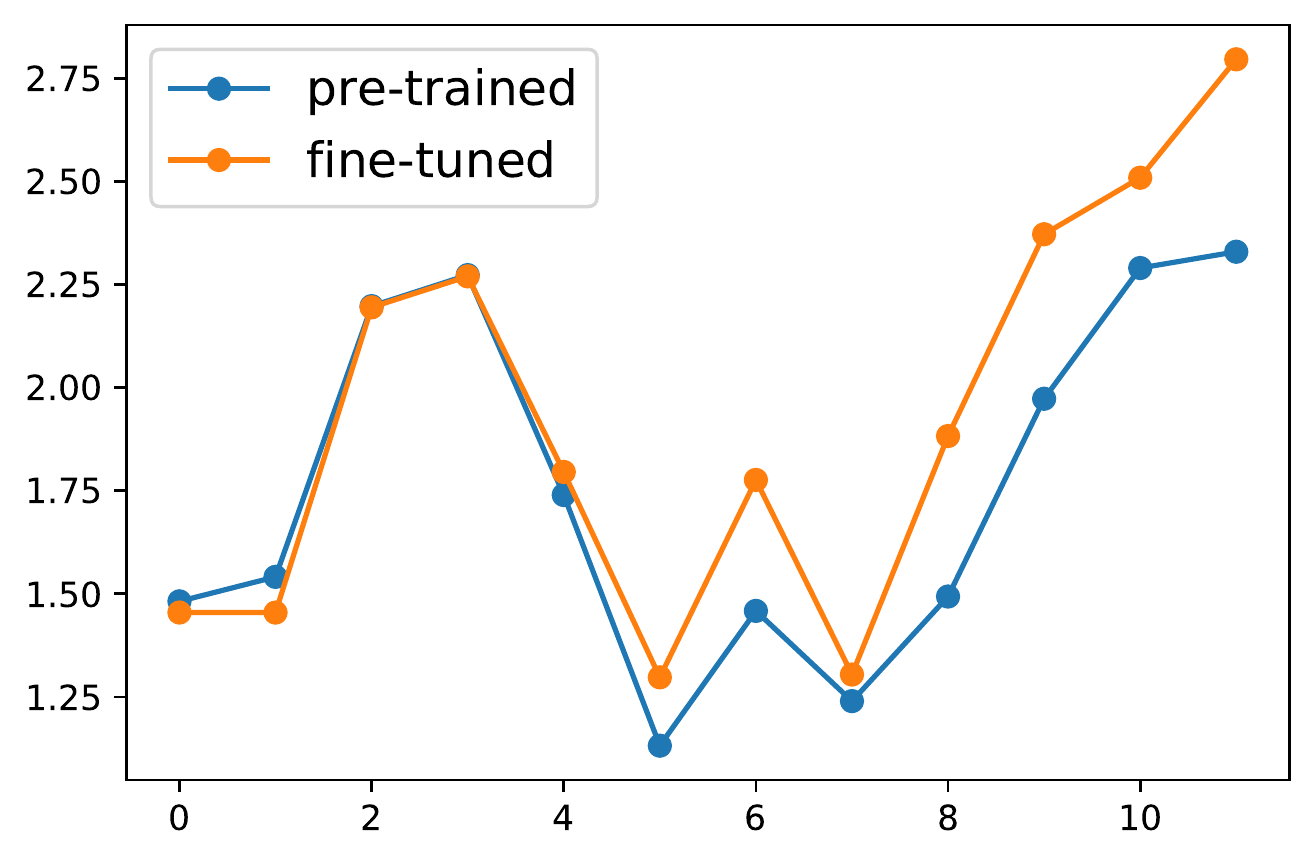}
        \caption{ICT}
        \label{fig:attn-ict}
    \end{subfigure}~
    \begin{subfigure}[t]{0.33\textwidth}
        \centering
        \includegraphics[width=\textwidth]{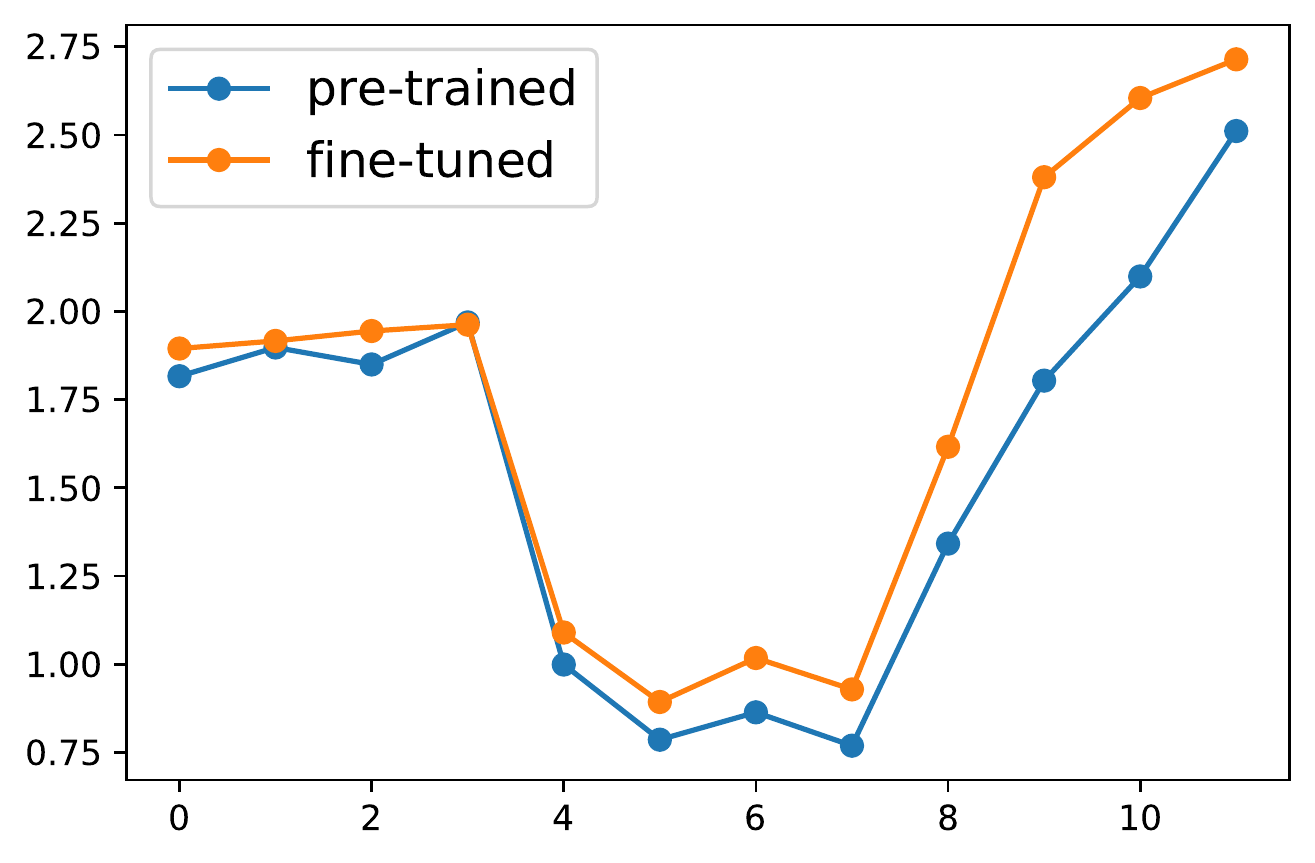}
        \caption{Condenser}
        \label{fig:attn-cd}
    \end{subfigure}%

\caption{Attention patterns in pre-trained v.s. fine-tuned BERT, ICT and Condenser.}
\label{fig:cls-attn}
\end{figure*}

\paragraph{Results} In \autoref{tab:marco}, we again find in low data, ICT and Condenser initialized retriever outperforms BERT by big margins. As it gets to 10K training data, 2\% of the full training set, all dense retrievers outperform BM25, with ICT and Condenser retaining their margin over BERT. Condenser can already show comparable performance in recall and NDCG to BERT trained on the full training set. We also observe that Condenser can outperform ICT at various train size, suggesting that the general LM pre-training of Condenser help it better generalize across domains than task-specific ICT.

\begin{table}[h]
\centering
\begin{tabular}{ l || c |  c }
\hline \hline
 & batch size & MRR@10 \\
\hline 
\textbf{RocketQA} &\\
Cross-batch  & 8192 & 0.333 \\
+ Hard negatives & 4096 & 0.260 \\
+ Denoise & 4096 & 0.364\\
+  Data augmentation & 4096 & \textbf{0.370}\\
\hline
\textbf{Condenser} & \\
w/o hard negatives & 64 &0.338  \\
w/ hard negatives & 64 & 0.366  \\
\hline
\hline 
\end{tabular}
\caption{Comparison with RocketQA MARCO Dev.}
\label{tab:rocket}
\end{table}

In \autoref{tab:marco-full}, we compare full train performance of various system. We see various training techniques help significantly improve over vanilla fine-tuning. Condenser can further outperform these models by big margins, showing the benefits brought by pre-training. Without involving complex training techniques, or making model/retrieval heavy, Condenser can already show slightly better performance than RocketQA.

We further give a comparison with RocketQA variants in \autoref{tab:rocket} to understand more costly strategies: very large batch, denoise hard negatives, and data augmentation. RocketQA authors find mined hard negatives contain false negatives detrimental to bi-encoder training as shown in the table and propose to use cross-encoder to relabel and denoise them, a process however thousands of times more costly than hard negative mining. They further employ a data augmentation technique, using a cross encoder to label external data. Here, we see Condenser trained with batch size 64 and BM25 negatives has better performance than RocketQA with 8192 batch size. More importantly, Condenser \emph{is able} to resist noise in mined hard negatives, getting a decent boost training with mined hard negatives, unlike RocketQA whose performance drops a lot without denoise. We see that Condenser removes the need for many sophisticated training techniques: it is only outperformed by the RocketQA variant that uses external data~(data augmentation).

Interestingly, our runs of BERT~(DPR) + HN have decent performance improvement over BERT in all retrieval tasks, sometimes better than active mining ANCE on both QA and Web Search. This contradicts the finding in RocketQA that directly mined hard negatives hurts performance. Recall our hard negatives are mined by Condenser retriever, which we conjecture has produced higher quality hard negatives. The finding suggests that mined hard negatives may not be retriever-dependent. There exist universally better ones, which can be found with a more effective retriever.

%% file: analysis.tex
\section{Attention Analysis}
\label{sec:analysis}
Condenser is built upon the idea that typical pre-trained LM lacks proper attention structure. We already see that we can fix the issue by pre-training with Condenser in the last section. In this section, we provide a more in-depth attention analysis: we compare attention behaviors among pre-trained/fine-tuned BERT, ICT, and Condenser. We use an analytical method proposed by \citet{Clark2019WhatDB}, characterizing the attention patterns of CLS by measuring its attention entropy. A higher entropy indicates broader attention and a lower more focused attention. Similar to \citet{Clark2019WhatDB}, we show CLS attention entropy at each layer, averaged over all heads, and averaged over 1k randomly picked Wikipedia sections. 

In \autoref{fig:cls-attn}, we plot attention from CLS of various models. We see in \autoref{fig:attn-bert} that BERT has a drastic change in attention pattern between pre-trained and fine-tuned models. This again confirmed our theory that typical Transformer Encoder LMs are not ready to be fined-tuned into bi-encoder, but need to go through big internal structural changes. In comparison, we see in Figures \ref{fig:attn-ict}, \ref{fig:attn-cd} that task-specific pre-trained ICT and LM pre-trained Condenser only have small changes, retaining general attention structure. In other words, ICT and Condenser both established structural readiness, but in very different ways. Both ICT and Condenser have broadening attention~(increased entropy) in the later layers, potentially because the actual search task requires aggregating more high-level concepts than pre-training. The results here again confirm our theory, that a ready-to-use structure can be easier to train; their structures only need small changes to work as an effective bi-encoder.

%% file: conclusion.tex
\section{Conclusion}
 Fine-tuning from a pre-trained LM initializer like BERT has become a very common practice in NLP. In this paper, we however question if models like BERT are the most proper initializer for bi-encoder. We find typical pre-trained LM does not have an internal attention structure ready for bi-encoder. They cannot effectively condense information into a single vector dense representation. We propose a new architecture, Condenser, which establishes readiness in structure with LM pre-training. We show Condenser is effective for a variety of tasks, sentence similarity, question answering retrieval, and web search retrieval. With low data, Condenser shows comparable performance to task-specific pre-trained models. It also provides a new pre-training perspective in learning effective retrievers than fine-tuning strategies. With sufficient training, Condenser and direct fine-tuning can be a lightweight alternative to many sophisticated training techniques. 

Positive results with Condenser show that structural readiness is a fundamental property in easy-to-train bi-encoders. Our attention analysis reveals both Condenser and task-specific pre-trained model establish structural readiness, suggesting task-specific objective may not be necessary.
Researchers can use this finding to guide the study of better LM for bi-encoder, for example, explore training Condenser with other LM objectives.

One big advantage of BERT is that after cumbersome pre-training for once, fine-tuning is easy with this universal model initializer. This is however not true for BERT bi-encoder, especially retriever, which needs careful and costly training. Condenser extends this benefit of BERT to bi-encoder. Practitioners on a limited budget can replace BERT  with our pre-trained Condenser as the initializer to get an instant performance boost. Meanwhile, for those aiming at the best performance, training techniques and Condenser can be combined. As we have demonstrated the combined effect of hard negatives and Condenser, sophisticated but better techniques can be further incorporated to train Condenser. 

%% file: appendix.tex
\section{Appendix}
\label{sec:appendix}
\subsection{Hyper Parameters Settings}
\paragraph{STS-b} The training follows hyper-parameter settings in \citet{reimers-gurevych-2019-sentence}, Adam optimizer, a learning rate of 2e-5 with linear schedule, and 4 epochs. For low data setup, we search best epoch number in \{4,8\} for BERT and apply those to all other pre-trained models.
\paragraph{Wikipedia Section Distinction} The training follows hyper-parameter settings in \citet{reimers-gurevych-2019-sentence}, Adam optimizer, a learning rate of 2e-5 with linear schedule and 1 epoch. For low data setup, we search best epoch number in \{1,4,8\} for BERT and apply those to all other pre-trained models.
\paragraph{Open QA} We follow hyperparameter settings in \citet{karpukhin-etal-2020-dense}, 128 batch size, 1 BM25 negative, in-batch negatives, 40 epochs, 1e-5 learning rate and linear schedule with warmup. Low data share the same setting as we found 40 epochs are enough for convergence.
\paragraph{Web Search} We train with Adam optimizer, learning rate of 5e-6 for 3 epochs with a total batch size of 64: 8 query $\times$ 8 passages. For low data setup, we search best epoch number in \{5, 10, 40\} for BERT and apply those to all other pre-trained models.
\subsection{Model Size}
In our experiments, Condenser during fine-tuning has the same number of parameters as BERT base, about 100 M. Adding the head during pre-training, there are roughly 120 M parameters. 

\subsection{ICT Model}
\label{app:ict}
Our ICT model comes from \citet{lee-etal-2019-latent}. It is trained with a batch size of 4096. ICT's effectiveness in low data setup was verified and thoroughly studied by \citet{chang2020pretraining}. \citet{chang2020pretraining} also introduces two other pre-training tasks Body First Selection and Wiki Link Prediction. They heavily depend on Wikipedia like structure and knowledge of the structure during pre-training and therefore does not apply in general situations. Meanwhile, adding them improves over ICT by only around 1\% and \citet{chang2020pretraining} has not released their model checkpoints. Therefore we chose to use the ICT checkpoint. 

Difficulties in reproducing these models come from the large batch requirement and the contrastive loss in ICT. Both \citet{lee-etal-2019-latent} and \citet{chang2020pretraining} find it critical to use large batch: \citet{lee-etal-2019-latent} uses a 4096 batch and \citet{chang2020pretraining} a 8192 batch. Both were trained with Google's cloud TPU. In comparison, our GPU can fit a batch of only 64. The contrastive loss uses the entire batch as the negative pool to learn the embedding space. Using gradient accumulation will reduce this pool size by several factors, leading to a bad pre-trained model. In comparison, our Condenser is based on instance-wise MLM loss and can naively use gradient accumulation.

We convert the original Tensorflow Checkpoint into Pytorch with huggingface conversion script. We don't use the linear projection layer that maps the 768 BERT embedding vector to 128 so that the embedding capacity is kept the same as retrievers in \citet{karpukhin-etal-2020-dense}.

\begin{table}[h]
\centering
\scalebox{0.9}{
\begin{tabular}{ l || c  | c }
\hline \hline
 & MS-MARCO Dev & DL2019 \\
Model & MRR@100 & NDCG@10 \\
\hline
BM25 & 0.230 & 0.519 \\
DeepCT & 0.320 & 0.544  \\
\hline
BERT & 0.340 & 0.546     \\
ME-BERT & n.a. & 0.588 \\
ANCE &  0.382 & \textbf{0.615}  \\
\hline
Condenser & 0.375 & 0.569 \\
Condenser + HN & \textbf{0.404} & 0.597 \\
\hline \hline 
\end{tabular}
}
\caption{Full train setup on MS-MARCO Document. Results not available are denoted `n.a.'}
\label{tab:marco-doc-full}
\vspace{-0.5cm}
\end{table}

\subsection{Document Retrieval}
\label{app:mdoc}
Recent works~\cite{xiong2021approximate,Luan2020SparseDA} explored retrieving long documents with the MS-MARCO document ranking dataset~\cite{bajaj2018ms}. There are several issues with this data set. The training set is not directly constructed but synthesizing from the passage ranking data set label. \citet{xiong2021approximate} find that the judgment in its TREC DL2019 test set biased towards BM25 and other lexical retrieval systems than dense retrievers. Meanwhile, \citet{Luan2020SparseDA} find single vector representation has a capacity issue in encoding long documents. To prevent these confounding from affecting our discussion, we opted to defer the experiment to this appendix. Here we use two query sets, MS-MARCO Document Dev and TREC DL2019. We report official metrics MRR@100 on Dev and NDCG@10 on DL2019. Results are recorded in \autoref{tab:marco-doc-full}. Condenser improves over BERT by a large margin and adding HN also boosts its performance. Condenser + HN performs the best on Dev. On the other hand, we see ANCE is the best on DL2019. We conjecture the reason is that use of BM25 negatives in many systems is not favorable towards DL2019 labels that favor lexical retrievers. The multi rounds of negative mining help ANCE get rid of the negative effect of BM25 negatives.

\subsection{Engineering Detail} 
We implement Condenser~(from BERT) in Pytorch~\cite{pytorch} based on the BERT implementation in huggingface transformers package~\cite{hf-transformers}. As our adjustments go only into the model architecture and the LM objective is kept unchanged, we only need to modify the modeling file and reuse the pre-training pipeline from huggingface.

\subsection{Link To Datasets}
\paragraph{Sentence Similarity} Cleaned up version can be found in the sentence transformer repo \url{https://github.com/UKPLab/sentence-transformers}.
\paragraph{Open QA} We use cleaned up open qa data from DPR \url{https://github.com/facebookresearch/DPR/}.
\paragraph{Web Search} MS-MARCO data can found on its homepage \url{https://microsoft.github.io/msmarco/}.

%% file: emnlp2021.bbl
\begin{thebibliography}{46}
\expandafter\ifx\csname natexlab\endcsname\relax\def\natexlab#1{#1}\fi

\bibitem[{Bajaj et~al.(2018)Bajaj, Campos, Craswell, Deng, Gao, Liu, Majumder,
  McNamara, Mitra, Nguyen, Rosenberg, Song, Stoica, Tiwary, and
  Wang}]{bajaj2018ms}
Payal Bajaj, Daniel Campos, Nick Craswell, Li~Deng, Jianfeng Gao, Xiaodong Liu,
  Rangan Majumder, Andrew McNamara, Bhaskar Mitra, Tri Nguyen, Mir Rosenberg,
  Xia Song, Alina Stoica, Saurabh Tiwary, and Tong Wang. 2018.
\newblock \href {http://arxiv.org/abs/1611.09268} {Ms marco: A human generated
  machine reading comprehension dataset}.

\bibitem[{Cer et~al.(2017)Cer, Diab, Agirre, Lopez-Gazpio, and
  Specia}]{cer-etal-2017-semeval}
Daniel Cer, Mona Diab, Eneko Agirre, I{\~n}igo Lopez-Gazpio, and Lucia Specia.
  2017.
\newblock \href {https://doi.org/10.18653/v1/S17-2001} {{S}em{E}val-2017 task
  1: Semantic textual similarity multilingual and crosslingual focused
  evaluation}.
\newblock In \emph{Proceedings of the 11th International Workshop on Semantic
  Evaluation ({S}em{E}val-2017)}, pages 1--14, Vancouver, Canada. Association
  for Computational Linguistics.

\bibitem[{Cer et~al.(2018)Cer, Yang, yi~Kong, Hua, Limtiaco, John, Constant,
  Guajardo-Cespedes, Yuan, Tar, Sung, Strope, and Kurzweil}]{cer2018universal}
Daniel Cer, Yinfei Yang, Sheng yi~Kong, Nan Hua, Nicole Limtiaco, Rhomni~St.
  John, Noah Constant, Mario Guajardo-Cespedes, Steve Yuan, Chris Tar,
  Yun-Hsuan Sung, Brian Strope, and Ray Kurzweil. 2018.
\newblock \href {http://arxiv.org/abs/1803.11175} {Universal sentence encoder}.

\bibitem[{Chang et~al.(2020)Chang, Yu, Chang, Yang, and
  Kumar}]{chang2020pretraining}
Wei-Cheng Chang, Felix~X. Yu, Yin-Wen Chang, Yiming Yang, and Sanjiv Kumar.
  2020.
\newblock \href {https://openreview.net/forum?id=rkg-mA4FDr} {Pre-training
  tasks for embedding-based large-scale retrieval}.
\newblock In \emph{International Conference on Learning Representations}.

\bibitem[{Chen et~al.(2020)Chen, Yang, Raman, Bendersky, Yeh, Zhou, Najork,
  Cai, and Emadzadeh}]{chen-etal-2020-dipair}
Jiecao Chen, Liu Yang, Karthik Raman, Michael Bendersky, Jung-Jung Yeh, Yun
  Zhou, Marc Najork, Danyang Cai, and Ehsan Emadzadeh. 2020.
\newblock \href {https://doi.org/10.18653/v1/2020.findings-emnlp.264}
  {{D}i{P}air: Fast and accurate distillation for trillion-scale text matching
  and pair modeling}.
\newblock In \emph{Findings of the Association for Computational Linguistics:
  EMNLP 2020}, pages 2925--2937, Online. Association for Computational
  Linguistics.

\bibitem[{Clark et~al.(2019)Clark, Khandelwal, Levy, and
  Manning}]{Clark2019WhatDB}
Kevin Clark, Urvashi Khandelwal, Omer Levy, and Christopher~D. Manning. 2019.
\newblock What does bert look at? an analysis of bert's attention.
\newblock \emph{ArXiv}, abs/1906.04341.

\bibitem[{Conneau and Kiela(2018)}]{Conneau2018SentEvalAE}
Alexis Conneau and Douwe Kiela. 2018.
\newblock Senteval: An evaluation toolkit for universal sentence
  representations.
\newblock \emph{ArXiv}, abs/1803.05449.

\bibitem[{Conneau et~al.(2017)Conneau, Kiela, Schwenk, Barrault, and
  Bordes}]{conneau-etal-2017-supervised}
Alexis Conneau, Douwe Kiela, Holger Schwenk, Lo{\"\i}c Barrault, and Antoine
  Bordes. 2017.
\newblock \href {https://doi.org/10.18653/v1/D17-1070} {Supervised learning of
  universal sentence representations from natural language inference data}.
\newblock In \emph{Proceedings of the 2017 Conference on Empirical Methods in
  Natural Language Processing}, pages 670--680, Copenhagen, Denmark.
  Association for Computational Linguistics.

\bibitem[{Dai and Callan(2019)}]{DeepCT}
Zhuyun Dai and J.~Callan. 2019.
\newblock Context-aware sentence/passage term importance estimation for first
  stage retrieval.
\newblock \emph{ArXiv}, abs/1910.10687.

\bibitem[{Dai et~al.(2020)Dai, Lai, Yang, and Le}]{Dai2020FunnelTransformerFO}
Zihang Dai, Guokun Lai, Yiming Yang, and Quoc~V. Le. 2020.
\newblock Funnel-transformer: Filtering out sequential redundancy for efficient
  language processing.
\newblock \emph{ArXiv}, abs/2006.03236.

\bibitem[{Devlin et~al.(2019)Devlin, Chang, Lee, and
  Toutanova}]{devlin-etal-2019-bert}
Jacob Devlin, Ming-Wei Chang, Kenton Lee, and Kristina Toutanova. 2019.
\newblock \href {https://doi.org/10.18653/v1/N19-1423} {{BERT}: Pre-training of
  deep bidirectional transformers for language understanding}.
\newblock In \emph{Proceedings of the 2019 Conference of the North {A}merican
  Chapter of the Association for Computational Linguistics: Human Language
  Technologies, Volume 1 (Long and Short Papers)}, pages 4171--4186,
  Minneapolis, Minnesota. Association for Computational Linguistics.

\bibitem[{Ein~Dor et~al.(2018)Ein~Dor, Mass, Halfon, Venezian, Shnayderman,
  Aharonov, and Slonim}]{ein-dor-etal-2018-learning}
Liat Ein~Dor, Yosi Mass, Alon Halfon, Elad Venezian, Ilya Shnayderman, Ranit
  Aharonov, and Noam Slonim. 2018.
\newblock \href {https://doi.org/10.18653/v1/P18-2009} {Learning thematic
  similarity metric from article sections using triplet networks}.
\newblock In \emph{Proceedings of the 56th Annual Meeting of the Association
  for Computational Linguistics (Volume 2: Short Papers)}, pages 49--54,
  Melbourne, Australia. Association for Computational Linguistics.

\bibitem[{Gao and Callan(2021)}]{gao2021unsupervised}
Luyu Gao and Jamie Callan. 2021.
\newblock \href {http://arxiv.org/abs/2108.05540} {Unsupervised corpus aware
  language model pre-training for dense passage retrieval}.

\bibitem[{Gao et~al.(2020)Gao, Dai, and Callan}]{gao-etal-2020-modularized}
Luyu Gao, Zhuyun Dai, and Jamie Callan. 2020.
\newblock \href {https://doi.org/10.18653/v1/2020.emnlp-main.342} {Modularized
  transfomer-based ranking framework}.
\newblock In \emph{Proceedings of the 2020 Conference on Empirical Methods in
  Natural Language Processing (EMNLP)}, pages 4180--4190, Online. Association
  for Computational Linguistics.

\bibitem[{Gao et~al.(2021{\natexlab{a}})Gao, Dai, and
  Callan}]{gao-etal-2021-coil}
Luyu Gao, Zhuyun Dai, and Jamie Callan. 2021{\natexlab{a}}.
\newblock \href {https://doi.org/10.18653/v1/2021.naacl-main.241} {{COIL}:
  Revisit exact lexical match in information retrieval with contextualized
  inverted list}.
\newblock In \emph{Proceedings of the 2021 Conference of the North American
  Chapter of the Association for Computational Linguistics: Human Language
  Technologies}, pages 3030--3042, Online. Association for Computational
  Linguistics.

\bibitem[{Gao et~al.(2021{\natexlab{b}})Gao, Dai, Chen, Fan, Durme, and
  Callan}]{CLEAR}
Luyu Gao, Zhuyun Dai, Tongfei Chen, Zhen Fan, Benjamin~Van Durme, and Jamie
  Callan. 2021{\natexlab{b}}.
\newblock Complement lexical retrieval model with semantic residual embeddings.
\newblock In \emph{Advances in Information Retrieval - 43rd European Conference
  on {IR} Research, {ECIR} 2021, Virtual Event, March 28 - April 1, 2021,
  Proceedings, Part {I}}.

\bibitem[{Gao et~al.(2021{\natexlab{c}})Gao, Zhang, Han, and
  Callan}]{gao-etal-2021-scaling}
Luyu Gao, Yunyi Zhang, Jiawei Han, and Jamie Callan. 2021{\natexlab{c}}.
\newblock \href {https://doi.org/10.18653/v1/2021.repl4nlp-1.31} {Scaling deep
  contrastive learning batch size under memory limited setup}.
\newblock In \emph{Proceedings of the 6th Workshop on Representation Learning
  for NLP (RepL4NLP-2021)}, pages 316--321, Online. Association for
  Computational Linguistics.

\bibitem[{Guo et~al.(2020)Guo, Sun, Lindgren, Geng, Simcha, Chern, and
  Kumar}]{avq_2020}
Ruiqi Guo, Philip Sun, Erik Lindgren, Quan Geng, David Simcha, Felix Chern, and
  Sanjiv Kumar. 2020.
\newblock \href {https://arxiv.org/abs/1908.10396} {Accelerating large-scale
  inference with anisotropic vector quantization}.
\newblock In \emph{International Conference on Machine Learning}.

\bibitem[{Guu et~al.(2020)Guu, Lee, Tung, Pasupat, and Chang}]{Guu2020REALMRL}
Kelvin Guu, Kenton Lee, Z.~Tung, Panupong Pasupat, and Ming-Wei Chang. 2020.
\newblock Realm: Retrieval-augmented language model pre-training.
\newblock \emph{ArXiv}, abs/2002.08909.

\bibitem[{Houlsby et~al.(2019)Houlsby, Giurgiu, Jastrzebski, Morrone,
  de~Laroussilhe, Gesmundo, Attariyan, and
  Gelly}]{Houlsby2019ParameterEfficientTL}
N.~Houlsby, A.~Giurgiu, Stanislaw Jastrzebski, Bruna Morrone, Quentin
  de~Laroussilhe, Andrea Gesmundo, Mona Attariyan, and S.~Gelly. 2019.
\newblock Parameter-efficient transfer learning for nlp.
\newblock In \emph{ICML}.

\bibitem[{Howard and Ruder(2018)}]{howard-ruder-2018-universal}
Jeremy Howard and Sebastian Ruder. 2018.
\newblock \href {https://doi.org/10.18653/v1/P18-1031} {Universal language
  model fine-tuning for text classification}.
\newblock In \emph{Proceedings of the 56th Annual Meeting of the Association
  for Computational Linguistics (Volume 1: Long Papers)}, pages 328--339,
  Melbourne, Australia. Association for Computational Linguistics.

\bibitem[{Johnson et~al.(2017)Johnson, Douze, and J{\'e}gou}]{JDH17}
Jeff Johnson, Matthijs Douze, and Herv{\'e} J{\'e}gou. 2017.
\newblock Billion-scale similarity search with gpus.
\newblock \emph{arXiv preprint arXiv:1702.08734}.

\bibitem[{Joshi et~al.(2017)Joshi, Choi, Weld, and
  Zettlemoyer}]{joshi-etal-2017-triviaqa}
Mandar Joshi, Eunsol Choi, Daniel Weld, and Luke Zettlemoyer. 2017.
\newblock \href {https://doi.org/10.18653/v1/P17-1147} {{T}rivia{QA}: A large
  scale distantly supervised challenge dataset for reading comprehension}.
\newblock In \emph{Proceedings of the 55th Annual Meeting of the Association
  for Computational Linguistics (Volume 1: Long Papers)}, pages 1601--1611,
  Vancouver, Canada. Association for Computational Linguistics.

\bibitem[{Karpukhin et~al.(2020)Karpukhin, Oguz, Min, Lewis, Wu, Edunov, Chen,
  and Yih}]{karpukhin-etal-2020-dense}
Vladimir Karpukhin, Barlas Oguz, Sewon Min, Patrick Lewis, Ledell Wu, Sergey
  Edunov, Danqi Chen, and Wen-tau Yih. 2020.
\newblock \href {https://doi.org/10.18653/v1/2020.emnlp-main.550} {Dense
  passage retrieval for open-domain question answering}.
\newblock In \emph{Proceedings of the 2020 Conference on Empirical Methods in
  Natural Language Processing (EMNLP)}, pages 6769--6781, Online. Association
  for Computational Linguistics.

\bibitem[{Kiros et~al.(2015)Kiros, Zhu, Salakhutdinov, Zemel, Torralba,
  Urtasun, and Fidler}]{kiros2015skipthought}
Ryan Kiros, Yukun Zhu, Ruslan Salakhutdinov, Richard~S. Zemel, Antonio
  Torralba, Raquel Urtasun, and Sanja Fidler. 2015.
\newblock \href {http://arxiv.org/abs/1506.06726} {Skip-thought vectors}.

\bibitem[{Kwiatkowski et~al.(2019)Kwiatkowski, Palomaki, Redfield, Collins,
  Parikh, Alberti, Epstein, Polosukhin, Devlin, Lee, Toutanova, Jones, Kelcey,
  Chang, Dai, Uszkoreit, Le, and Petrov}]{kwiatkowski-etal-2019-natural}
Tom Kwiatkowski, Jennimaria Palomaki, Olivia Redfield, Michael Collins, Ankur
  Parikh, Chris Alberti, Danielle Epstein, Illia Polosukhin, Jacob Devlin,
  Kenton Lee, Kristina Toutanova, Llion Jones, Matthew Kelcey, Ming-Wei Chang,
  Andrew~M. Dai, Jakob Uszkoreit, Quoc Le, and Slav Petrov. 2019.
\newblock \href {https://doi.org/10.1162/tacl_a_00276} {Natural questions: A
  benchmark for question answering research}.
\newblock \emph{Transactions of the Association for Computational Linguistics},
  7:452--466.

\bibitem[{Lan et~al.(2020)Lan, Chen, Goodman, Gimpel, Sharma, and
  Soricut}]{Lan2020ALBERTAL}
Zhenzhong Lan, Mingda Chen, Sebastian Goodman, Kevin Gimpel, Piyush Sharma, and
  Radu Soricut. 2020.
\newblock Albert: A lite bert for self-supervised learning of language
  representations.
\newblock \emph{ArXiv}, abs/1909.11942.

\bibitem[{Lee et~al.(2019)Lee, Chang, and Toutanova}]{lee-etal-2019-latent}
Kenton Lee, Ming-Wei Chang, and Kristina Toutanova. 2019.
\newblock \href {https://doi.org/10.18653/v1/P19-1612} {Latent retrieval for
  weakly supervised open domain question answering}.
\newblock In \emph{Proceedings of the 57th Annual Meeting of the Association
  for Computational Linguistics}, pages 6086--6096, Florence, Italy.
  Association for Computational Linguistics.

\bibitem[{Lewis et~al.(2020)Lewis, Liu, Goyal, Ghazvininejad, Mohamed, Levy,
  Stoyanov, and Zettlemoyer}]{lewis-etal-2020-bart}
Mike Lewis, Yinhan Liu, Naman Goyal, Marjan Ghazvininejad, Abdelrahman Mohamed,
  Omer Levy, Veselin Stoyanov, and Luke Zettlemoyer. 2020.
\newblock \href {https://doi.org/10.18653/v1/2020.acl-main.703} {{BART}:
  Denoising sequence-to-sequence pre-training for natural language generation,
  translation, and comprehension}.
\newblock In \emph{Proceedings of the 58th Annual Meeting of the Association
  for Computational Linguistics}, pages 7871--7880, Online. Association for
  Computational Linguistics.

\bibitem[{Lin et~al.(2020)Lin, Yang, and Lin}]{Lin2020DistillingDR}
Sheng-Chieh Lin, Jheng-Hong Yang, and Jimmy Lin. 2020.
\newblock Distilling dense representations for ranking using tightly-coupled
  teachers.
\newblock \emph{ArXiv}, abs/2010.11386.

\bibitem[{Liu et~al.(2019)Liu, Ott, Goyal, Du, Joshi, Chen, Levy, Lewis,
  Zettlemoyer, and Stoyanov}]{Liu2019RoBERTaAR}
Y.~Liu, Myle Ott, Naman Goyal, Jingfei Du, Mandar Joshi, Danqi Chen, Omer Levy,
  M.~Lewis, Luke Zettlemoyer, and Veselin Stoyanov. 2019.
\newblock Roberta: A robustly optimized bert pretraining approach.
\newblock \emph{ArXiv}, abs/1907.11692.

\bibitem[{Luan et~al.(2020)Luan, Eisenstein, Toutanova, and
  Collins}]{Luan2020SparseDA}
Y.~Luan, Jacob Eisenstein, Kristina Toutanova, and Michael Collins. 2020.
\newblock Sparse, dense, and attentional representations for text retrieval.
\newblock \emph{ArXiv}, abs/2005.00181.

\bibitem[{MacAvaney et~al.(2020)MacAvaney, Nardini, Perego, Tonellotto,
  Goharian, and Frieder}]{MacAvaney2020EfficientDR}
Sean MacAvaney, F.~Nardini, R.~Perego, N.~Tonellotto, Nazli Goharian, and
  O.~Frieder. 2020.
\newblock Efficient document re-ranking for transformers by precomputing term
  representations.
\newblock \emph{Proceedings of the 43rd International ACM SIGIR Conference on
  Research and Development in Information Retrieval}.

\bibitem[{Mao et~al.(2020)Mao, He, Liu, Shen, Gao, Han, and
  Chen}]{mao2020generationaugmented}
Yuning Mao, Pengcheng He, Xiaodong Liu, Yelong Shen, Jianfeng Gao, Jiawei Han,
  and Weizhu Chen. 2020.
\newblock \href {http://arxiv.org/abs/2009.08553} {Generation-augmented
  retrieval for open-domain question answering}.

\bibitem[{Nogueira and Lin(2019)}]{docTTTTTquery}
Rodrigo Nogueira and Jimmy Lin. 2019.
\newblock From doc2query to doctttttquery.

\bibitem[{Paszke et~al.(2019)Paszke, Gross, Massa, Lerer, Bradbury, Chanan,
  Killeen, Lin, Gimelshein, Antiga, Desmaison, Kopf, Yang, DeVito, Raison,
  Tejani, Chilamkurthy, Steiner, Fang, Bai, and Chintala}]{pytorch}
Adam Paszke, Sam Gross, Francisco Massa, Adam Lerer, James Bradbury, Gregory
  Chanan, Trevor Killeen, Zeming Lin, Natalia Gimelshein, Luca Antiga, Alban
  Desmaison, Andreas Kopf, Edward Yang, Zachary DeVito, Martin Raison, Alykhan
  Tejani, Sasank Chilamkurthy, Benoit Steiner, Lu~Fang, Junjie Bai, and Soumith
  Chintala. 2019.
\newblock \href
  {http://papers.neurips.cc/paper/9015-pytorch-an-imperative-style-high-performance-deep-learning-library.pdf}
  {Pytorch: An imperative style, high-performance deep learning library}.
\newblock In H.~Wallach, H.~Larochelle, A.~Beygelzimer, F.~d\textquotesingle
  Alch\'{e}-Buc, E.~Fox, and R.~Garnett, editors, \emph{Advances in Neural
  Information Processing Systems 32}. Curran Associates, Inc.

\bibitem[{Peters et~al.(2018)Peters, Neumann, Iyyer, Gardner, Clark, Lee, and
  Zettlemoyer}]{peters-etal-2018-deep}
Matthew Peters, Mark Neumann, Mohit Iyyer, Matt Gardner, Christopher Clark,
  Kenton Lee, and Luke Zettlemoyer. 2018.
\newblock \href {https://doi.org/10.18653/v1/N18-1202} {Deep contextualized
  word representations}.
\newblock In \emph{Proceedings of the 2018 Conference of the North {A}merican
  Chapter of the Association for Computational Linguistics: Human Language
  Technologies, Volume 1 (Long Papers)}, pages 2227--2237, New Orleans,
  Louisiana. Association for Computational Linguistics.

\bibitem[{Qu et~al.(2020)Qu, Ding, Liu, Liu, Ren, Zhao, Dong, Wu, and
  Wang}]{Qu2020RocketQAAO}
Y.~Qu, Yuchen Ding, Jing Liu, Kai Liu, Ruiyang Ren, X.~Zhao, Daxiang Dong, Hua
  Wu, and H.~Wang. 2020.
\newblock Rocketqa: An optimized training approach to dense passage retrieval
  for open-domain question answering.
\newblock \emph{ArXiv}, abs/2010.08191.

\bibitem[{Reimers and Gurevych(2019)}]{reimers-gurevych-2019-sentence}
Nils Reimers and Iryna Gurevych. 2019.
\newblock \href {https://doi.org/10.18653/v1/D19-1410} {Sentence-{BERT}:
  Sentence embeddings using {S}iamese {BERT}-networks}.
\newblock In \emph{Proceedings of the 2019 Conference on Empirical Methods in
  Natural Language Processing and the 9th International Joint Conference on
  Natural Language Processing (EMNLP-IJCNLP)}, pages 3982--3992, Hong Kong,
  China. Association for Computational Linguistics.

\bibitem[{Ronneberger et~al.(2015)Ronneberger, Fischer, and
  Brox}]{RonnebergerFB15}
Olaf Ronneberger, Philipp Fischer, and Thomas Brox. 2015.
\newblock \href {https://doi.org/10.1007/978-3-319-24574-4\_28} {U-net:
  Convolutional networks for biomedical image segmentation}.
\newblock In \emph{Medical Image Computing and Computer-Assisted Intervention -
  {MICCAI} 2015 - 18th International Conference Munich, Germany, October 5 - 9,
  2015, Proceedings, Part {III}}, volume 9351 of \emph{Lecture Notes in
  Computer Science}, pages 234--241. Springer.

\bibitem[{Thakur et~al.(2020)Thakur, Reimers, Daxenberger, and
  Gurevych}]{thakur2020augmented}
Nandan Thakur, Nils Reimers, Johannes Daxenberger, and Iryna Gurevych. 2020.
\newblock \href {http://arxiv.org/abs/2010.08240} {Augmented sbert: Data
  augmentation method for improving bi-encoders for pairwise sentence scoring
  tasks}.

\bibitem[{Vaswani et~al.(2017)Vaswani, Shazeer, Parmar, Uszkoreit, Jones,
  Gomez, Kaiser, and Polosukhin}]{Vaswani2017AttentionIA}
Ashish Vaswani, Noam Shazeer, Niki Parmar, Jakob Uszkoreit, Llion Jones,
  Aidan~N. Gomez, L.~Kaiser, and Illia Polosukhin. 2017.
\newblock Attention is all you need.
\newblock \emph{ArXiv}, abs/1706.03762.

\bibitem[{Wang et~al.(2018)Wang, Singh, Michael, Hill, Levy, and
  Bowman}]{wang-etal-2018-glue}
Alex Wang, Amanpreet Singh, Julian Michael, Felix Hill, Omer Levy, and Samuel
  Bowman. 2018.
\newblock \href {https://doi.org/10.18653/v1/W18-5446} {{GLUE}: A multi-task
  benchmark and analysis platform for natural language understanding}.
\newblock In \emph{Proceedings of the 2018 {EMNLP} Workshop {B}lackbox{NLP}:
  Analyzing and Interpreting Neural Networks for {NLP}}, pages 353--355,
  Brussels, Belgium. Association for Computational Linguistics.

\bibitem[{Wolf et~al.(2019)Wolf, Debut, Sanh, Chaumond, Delangue, Moi, Cistac,
  Rault, Louf, Funtowicz, Davison, Shleifer, von Platen, Ma, Jernite, Plu, Xu,
  Scao, Gugger, Drame, Lhoest, and Rush}]{hf-transformers}
Thomas Wolf, Lysandre Debut, Victor Sanh, Julien Chaumond, Clement Delangue,
  Anthony Moi, Pierric Cistac, Tim Rault, Rémi Louf, Morgan Funtowicz, Joe
  Davison, Sam Shleifer, Patrick von Platen, Clara Ma, Yacine Jernite, Julien
  Plu, Canwen Xu, Teven~Le Scao, Sylvain Gugger, Mariama Drame, Quentin Lhoest,
  and Alexander~M. Rush. 2019.
\newblock Huggingface's transformers: State-of-the-art natural language
  processing.
\newblock \emph{ArXiv}, abs/1910.03771.

\bibitem[{Xiong et~al.(2021)Xiong, Xiong, Li, Tang, Liu, Bennett, Ahmed, and
  Overwijk}]{xiong2021approximate}
Lee Xiong, Chenyan Xiong, Ye~Li, Kwok-Fung Tang, Jialin Liu, Paul~N. Bennett,
  Junaid Ahmed, and Arnold Overwijk. 2021.
\newblock \href {https://openreview.net/forum?id=zeFrfgyZln} {Approximate
  nearest neighbor negative contrastive learning for dense text retrieval}.
\newblock In \emph{International Conference on Learning Representations}.

\bibitem[{Yang et~al.(2019)Yang, Dai, Yang, Carbonell, Salakhutdinov, and
  Le}]{Yang2019XLNetGA}
Z.~Yang, Zihang Dai, Yiming Yang, J.~Carbonell, R.~Salakhutdinov, and Quoc~V.
  Le. 2019.
\newblock Xlnet: Generalized autoregressive pretraining for language
  understanding.
\newblock In \emph{NeurIPS}.

\end{thebibliography}
